\documentclass[sigconf,screen]{acmart}

\AtBeginDocument{%
  \providecommand\BibTeX{{%
    \normalfont B\kern-0.5em{\scshape i\kern-0.25em b}\kern-0.8em\TeX}}}

\usepackage{colortbl}

\usepackage{multirow}
\usepackage{booktabs}

\usepackage{enumitem}
\usepackage{xspace}

\setcopyright{acmlicensed}
\acmPrice{15.00}
\acmDOI{10.1145/3533767.3534375}
\acmYear{2022}
\copyrightyear{2022}
\acmSubmissionID{issta22main-p78-p}
\acmISBN{978-1-4503-9379-9/22/07}
\acmConference[ISSTA '22]{Proceedings of the 31st ACM SIGSOFT International Symposium on Software Testing and Analysis}{July 18--22, 2022}{Virtual, South Korea}
\acmBooktitle{Proceedings of the 31st ACM SIGSOFT International Symposium on Software Testing and Analysis (ISSTA '22), July 18--22, 2022, Virtual, South Korea}

\usepackage{tcolorbox}

\begin{document}

\title{Simple Techniques Work Surprisingly Well for Neural Network Test Prioritization and Active Learning (Replicability Study)}

\author{Michael Weiss}
\email{michael.weiss@usi.ch}
\orcid{0000-0002-8944-389X}
\affiliation{%
  \institution{Università della Svizzera italiana}
  \city{Lugano}
  \country{Switzerland}
  \postcode{6900}
}
\author{Paolo Tonella}
\email{paolo.tonella@usi.ch}
\orcid{0000-0003-3088-0339}
\affiliation{%
  \institution{Università della Svizzera italiana}
  \city{Lugano}
  \country{Switzerland}
  \postcode{6900}
}



\newcommand{\replipkg}[0]{reproduction package}
\newcommand{\replicatedPaper}[0]{Feng. et. al.~\cite{Feng2020deepgini}\xspace}

\begin{abstract}

Test Input Prioritizers (TIP) for Deep Neural Networks (DNN) are an important technique to handle the typically very large test datasets efficiently, saving computation and labelling costs. 
This is particularly true for large scale, deployed systems, where inputs observed in production are recorded to serve as potential test or training data for next versions of the system.
Feng et. al. propose DeepGini, a very fast and simple TIP and show that it outperforms more elaborate techniques such as neuron- and surprise coverage.
In a large-scale study (4 case studies, 8 test datasets, 32'200 trained models) we verify their findings.
However, we also find that other comparable or even simpler baselines from the field of uncertainty quantification, such as the predicted softmax likelihood or the entropy of the predicted softmax likelihoods perform equally well as DeepGini.

\end{abstract}

\maketitle

\section{Introduction}
\label{sec:intro}

\begin{figure}[b]
    \centering
    \fbox{
        \includegraphics[width=0.6\linewidth]{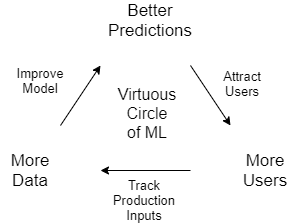}
    }
    \caption{Virtuous Circle of Machine Learning %
    }
    \label{fig:virt_circle_of_ml}
\end{figure}

\emph{Deep Neural Networks} (DNN) are typically trained and tested on large data sets.
\emph{Test input prioritizers} (TIP) allow developers to order or to identify a subset of the test inputs which -- similar to traditional software testing~\cite{ElbaumMR00,LiangER18,RothermelUCH01} -- detect faults at a reduced time and energy cost. 
In addition, and arguably even more importantly, TIP can be used with production ML (Machine Learning) based systems, to facilitate iterative improvements of the deployed model:
As nicely described through the \emph{Virtuous Circle of ML}~\cite{NgVirtuousCircle}, depicted in \autoref{fig:virt_circle_of_ml}, modern ML based systems profit from economies of scale, as the more often a ML based system is used, the more real-world production data it observes and collects, which in turn can be used to further test and improve its ML components, leading to better predictions and thus attracting even more users.
However, this iterative process  becomes quickly infeasible as  large scales are reached: 
With a very large number of users, observed production data become hard if not impossible to process for model improvement. 
TIP can be used to identify the small share of production inputs which expose model faults and should thus be integrated in a test suite, or which are generally insufficiently represented in the  training data and should thus be used for active learning, i.e., should be added to the training set of future training runs.
This is in line with the industrial practice of data-centric quality improvement, described by Andrej Karpathy, Tesla's director for AI, in a recent presentation~\cite{Karpathy2021TeslaSelfDrivingCVPR}.

In order to handle inputs of large scale systems, TIP must scale very well  as well. 
To this aim, Feng et. al. ~\cite{Feng2020deepgini} proposed \emph{DeepGini}, an approach for classification DNNs which assigns priority scores based on the softmax outputs of a DNN at negligible cost. 
The authors showed that DeepGini reliably outperformed classical DNN TIP from the two families of neuron coverage and surprise coverage.
In this paper, we replicate, extend and contextualize the findings obtained in that paper. 
Specifically, we make the following contributions:
\begin{itemize}[noitemsep,leftmargin=*]
\item \textbf{Replication} We successfully replicate the key findings by Feng et. al. ~\cite{Feng2020deepgini}, showing that DeepGini outperforms various types of surprise and neuron coverage metrics on execution time, as well as test prioritisation and active learning effectiveness.
\item \textbf{Comparison} We extended the list of TIP approaches compared against DeepGini with recent improved variants of surprise coverage and uncertainty quantifiers. 
Our experiments show that  uncertainty quantifiers, including a simple baseline that uses just the predicted class likelihood, perform comparably well.
\item \textbf{Statistical Evaluation} Previous work shows evidence that some TIP performances are highly sensitive to random influences during model training~\cite{Weiss2021-SA}. 
We thus perform our experiments on 100 individually trained models and discuss the statistical significance of our results. Such statistical analysis was not carried out in the original DeepGini paper~\cite{Feng2020deepgini}.
\item \textbf{Versatile Artifacts} Not only do we release all our code, as generally requested for replication studies~\cite{ISSTA2022CfP}, but we also publish  core components of the investigated techniques, which we think could be useful for a wide range of possible follow-up studies.
Our implementations of neuron coverage TIP, surprise coverage TIP and DeepGini will are easily installable through the \textsc{pypi} package manager. Our generated datasets are made available as standalone artifacts.
\end{itemize}
\section{Motivation}
\label{sec:motivation}
In the following, we motivate the use of DeepGini and similar TIP techniques in the test prioritization and active learning scenarios.

\subsection{Test Prioritization}
Test suites for traditional (non ML-based) software systems can become very large and running the full test suite might come at a large cost, both regarding time (making the development process inefficient) and energy consumption.
Correspondingly, TIP for traditional software has been investigated to give priority to tests which are more likely to fail, providing fast feedback about the discovered problems~\cite{ElbaumMR00,LiangER18,RothermelUCH01}.
When testing ML based systems, in addition to saving computation time and providing early feedback,
TIP is also important to mitigate the high cost associated with test data labelling. For instance, test data gathered from the production environment is typically unlabelled. Another example is automatically generated inputs, e.g. produced by fuzzing techniques, where clearly, the main cost does not come with  input generation, but from  manual labelling of the relevant inputs.

For companies adopting the virtuous circle of ML (see~\autoref{fig:virt_circle_of_ml}), TIP is extremely useful, because a constant stream of potentially useful test data is observed on deployed systems and collected for the continuous improvement of the model.
Here, not all inputs which lead to a \textit{model failure} (i.e., a wrong or inaccurate model prediction) lead to an observable \textit{system failure}, as the production system may be able to compensate for occasional deficiencies of the ML components.
Hence, it is nontrivial to identify production time inputs which should be collected and reported to the testing team. 

\subsection{Active Learning}
Active Learning builds on the idea that novel data, observed in production, can be used when training the next generation of the same model.
Clearly, before production data can be used, it has to be (usually manually) labelled.
Again, it becomes very important to select unlabelled raw data which has the potential to be most useful for  future model improvement. This is typically data which is \emph{out-of-distribution (OOD)} with respect to the previous training dataset. TIP could  drastically reduce the labelling cost by giving higher priority to OOD data, making active learning a viable activity in the virtuous circle of ML. 

Several recent papers target the problem of identifying the most relevant among the available raw inputs \cite{Berend2020cats, ma2021test, Kim2018, Feng2020deepgini, Stocco2020towards} to improve the behaviour of an ML based system by retraining its DNNs. TIP is clearly a very suitable candidate for such a task~\cite{Feng2020deepgini}.
Specifically, a TIP would be used as \emph{acquisition function}~\cite{gal2017acquisitionFunction} to prioritize unlabelled data according to their potential to improve a DNN's performance when such data is added to the training set. The selected, highest prioritized data would then be labelled (typically manually, by a human) and added to the training set for retraining.
Here, we note that the goal when using a TIP for active learning is now slightly shifted with respect to test prioritization:
We are not primarily interested in whether an input leads to a model failure (e.g., a mis-classification), but whether an input is out-of-distribution, i.e., insufficiently represented during training.

\newcommand{\BigO}[1]{$\mathcal{O}(#1)$}
\newcommand{\scipy}[0]{\textsc{SCIPY}}

\section{Approaches}
\label{sec:approaches}

In the following, we describe the different TIP techniques we evaluate in this paper.
For all of them, we use the following common notation: 
We let $n$ denote the number of training samples, $a$ the number of neurons considered by the specific approach and $k$ as a further model specific parameter.

\subsection{Neuron Coverage}
\label{sec:approach_neuron_coverage}

Similar to code coverage, Neuron Coverage (NC) can be used to prioritize a list of test inputs, either by descending amount of absolute coverage achieved by each test or by the additional amount of coverage that each test brings with respect to the overall coverage reached by previously executed tests.
NC creates a (boolean) coverage profile for every test input, where, individually, coverage of each DNN neuron is evaluated based on its activation. 
Depending on the specific variant, the coverage profile may distinguish different activation segments of each neuron and treat them as separate coverage targets. 
We refer to \replicatedPaper for a nice overview of different NC technqiues, and only provide a very short description of each approach in this section.

In \emph{Neuron Activation Coverage (NAC-$k$)}~\cite{Pei2017deepxplore}, a node is considered covered, if its activation is higher than $k$. 
To also take low activation into account, \emph{$k$-Multisection NC (KMNC-$k$)}~\cite{Ma2018deepgauge} the range of activations observed on the training set for each neuron is uniformly divided into $k$ segments, which are considered covered if the DNN neuron activations for the considered test inputs fall within each segment.
Taking the opposite direction, \emph{Neuron Boundary Coverage (NBC-$k$)}~\cite{Ma2018deepgauge} regards only two segments per node, representing activations that fall below or above the activation range (boundaries) observed during training. 
To  move further away from these boundaries and  detect novel, distant activations, the training boundaries can be shifted apart by $k\sigma$, where $\sigma$ denotes the standard deviations of the activations observed for the corresponding node on the training set.
As a special case of NBC-$k$, \emph{Strong Neuron Activation Coverage (SNAC-k)}~\cite{Ma2018deepgauge} considers only the upper bound, thus halving the size of the coverage profile.
Instead, \emph{Top-$k$ NC (TKNC-$k$)}~\cite{Ma2018deepgauge} considers a neuron covered if it belongs to the $k$ neurons with the highest activations within the same layer. 

\paragraph{CTM vs. CAM} 
Given a coverage profile, the coverage of a test input can be calculated as the percentage of \emph{true}'s in its coverage profile.
Prioritization of tests by decreasing coverage is denoted as \emph{Coverage-Total Method (CTM)}.
While CTM is fast and easy to run, it has the disadvantage that it does not necessarily aim for diversity in the prioritized tests, as different tests with high coverage might cover the same neurons, while tests that cover previously uncovered neurons might have a low total coverage, which would schedule them late in the execution.
To address this issue, \emph{Coverage-Additional Method (CAM)} aims to reach an overall combined coverage as quickly as possible, using a greedy approach. By starting with the test with  highest coverage, CAM then always adds the test which covers the most previously uncovered targets in the coverage profile. 
It thus automatically increases the diversity in the prioritized test list. 
The main disadvantage of CAM is its computational complexity, running in a time quadratic in the test set size.  

\paragraph{Our implementation}
The  primary computational challenge arou\-nd NC arises from memory requirements, which grow linear in $a$ for all NC and thus quickly become intractable when handling very large DNNs.
We mitigate this problem, to some extent, by using online batch processing: 
Collecting activation traces (which typically consist of 64 or 32 bit floats) and reducing them to the typically much smaller boolean coverage profiles in batches allow us to reduce memory consumption  by up to (almost) 64 times, depending on the size of the coverage profile of the used NC.

\subsection{Surprise Adequacy and -Coverage}
\label{sec:approach_surprise_adequacy}

Surprise Adequacy (SA) denotes a collection of techniques to measure how \emph{surprising} a DNN input is, i.e., how novel or out-of-distribution the DNN activations for a given input are with respect to the activations observed on the training set. 
While in theory SA measurements could be applied on the full set of neurons, similar to NC, in practice it is often limited to the activations of a single or few hidden layers. If we  denote such activation traces as $AT_{SA}$, we thus have $a = |AT_{SA}|$.
Considering only a subset of the DNN nodes when collecting $AT_{SA}$ brings two main advantages: 
First, later layers are responsible for processing higher level features extracted from the raw input data, making them more appropriate to measure surprise~\cite{Kim2018}.
Second, the use of fewer activations  allows for a faster and more memory efficient execution of SA.
The latter is particularly important, as the computational cost is a major shortcoming of most SA variants~\cite{Kim2020MDSA, Weiss2021-SA}.

\subsubsection{Likelihood-Based SA}
Likelihood-Based SA (LSA) estimates the negative log-likelihood of a test input's $AT_{SA}$ using a Gaussian Kernel Density Estimator (KDE) parameterized on the training set $AT_{SA}$.
Gaussian KDE is known to be slow to train, with  quadratic or even cubic runtimes in $n$~\cite{Wang2014KDE}. 
Moreover,  likelihood quantification at prediction time runs in \BigO{n \cdot m}, as inputs must be compared to all training data's activations~\cite{Kim2020MDSA}.
Fast implementations such as the one by \textsc{scikit-learn}\footnote{\url{https://docs.scipy.org/doc/scipy/reference/generated/scipy.stats.gaussian_kde.html}} mitigate this problem as long as $m$ and $n$ are not too large~\cite{Weiss2021-SA}.
To the best of our knowledge, all publicly available source code that accompanies papers using surprise adequacy \cite{Kim2018, Weiss2021-SA} rely on the \textsc{scikit-learn} implementation, which comes at a cost: 
The implementation is numerically unstable, and as we use it with many features, it is likely that an imprecise representation of the $AT_{SA}$ covariance matrix leads to a crash of the KDE\footnote{See e.g., \url{https://stackoverflow.com/a/66902455/}}.
Looking at the above mentioned previous implementations of LSA, this problem appears to have been mitigated by choosing a layer with only a few neurons, i.e., a low $a$, and by ignoring neurons whose activation's variance on the training set is below some threshold. 

\paragraph{Our implementation of LSA} 
Also relying on the fast and well tested \textsc{scikit-learn}'s \textit{GaussianKDE} implementation~\cite{scikit-learn}, our implementation aims to reduce the risk of KDE crashes due to numerical imprecision to a minimum:
Besides  providing a way to specify a threshold for minimum required variance in a neuron's activation trace, it also exposes an interface to specify a (relative or absolute) number of neurons to consider, which are selected by decreasing variance. 
In addition, we extended \textsc{scikit-learn}'s GaussianKDE to modify the covariance matrix of the training $AT_{SA}$ by adding a dynamically chosen, small $\epsilon$ to the covariances diagonal, making it positive semidefinite, while keeping the falsification of predictions small. 
Combined, these steps allow us to eliminate most crashes due to numerical instability.
In the very rare cases when GaussianKDE still fails, our implementation  fails gracefully, by returning a surprise value of 0. 

\subsubsection{Distance-Based SA}
Distance-Based Surprise Adequacy (DSA), which is defined only for classification problems, quantifies surprise as the ratio between a test input's $AT_{SA}$ (Euclidean) distance to the closest training $AT_{SA}$ of the same class, and the distance between that point and its closest training $AT_{SA}$ belonging to any other class. 
DSA has been shown to outperform LSA for test input prioritization, active learning \cite{Kim2018} and fault prediction~\cite{Weiss2021-SA}. 
The major disandvantage of DSA is clearly its computational cost~\cite{Weiss2021-SA}. 
For a single test input,  the distance to each $AT_{SA}$ has to be calculated,
leading to \BigO{a \cdot n} time and space complexities.
To mitigate this problem, only  a subset of the training $AT_{SA}$ has been considered, leading to acceptable DSA performance~\cite{Weiss2021-SA}. While this does not reduce the asymptotic complexity, it provides both speed and memory reductions proportionate to the selected sampling ratio.  
The same authors~\cite{Weiss2021-SA} also proposed to calculate DSA for multiple test inputs at the same time using multithreaded, vectorized batch calculations. 
While this leads to considerable practical improvements in terms of speed, it requires drastically more memory: 
With a batch size $b$ and $t$ threads, peak memory load can increase to \BigO{t \cdot b \cdot n \cdot m}, which can become problematic even for moderate batch and thread sizes. Let us consider for example a setting with $a=1600$ and $m=60,000$ (as in our MNIST case study). Choosing $t=8$ and $b=32$, and a float precision of 32 bits, would result in a worst-case peak memory load of more than 98.3GB.

\paragraph{Our implementation of DSA}
Based on the implementation by Weiss et. al.~\cite{Weiss2021-SA}, our implementation supports (uniform) training set subsampling and vectorized, multithreaded batched computation. 
Besides minor changes, we have replaced their batching and threading logic with one that reduces the variance in memory consumption, which thus allows for a much better tuning of the batch size $b$ for optimal performance. 

\subsubsection{Mahalanobis-Distance Based SA}
\emph{Mahalanobis-Distance Based SA (MDSA)}, which -- despite its name -- can be understood as an alternative to \emph{likelihood} based SA, aims to provide a  much more efficient way to compute the likelihood of a test's $AT_{SA}$, given the observed $AT_{SA}$ of the training set.
By requiring only the covariance matrix of the training set $AT_{SA}$, 
the Mahalanobis distance~\cite{Mahalanobis1936} can calculate the likelihood for the given input's $AT_{SA}$ efficiently, making 
prediction runtimes constant in the training set size.
Memory-wise, storing the covariance matrix costs \BigO{a^2}. 
To calculate the covariance matrix, the full set of training $AT_{SA}$ has to be collected (thus typically resulting in a setup runtime and memory requirement linear in $n$). 
In practice, similar to DSA and LSA, these setup requirements could potentially be relaxed by choosing a subset of training samples (similar to DSA) or only high-variance features (similar to LSA). 
Even more so, the memory requirements could be made constant in $n$ by using an  online covariance estimation algorithm, specifically designed to scale well for large datasets by estimating the covariance on a non-persistent stream of data ~\cite{Bennett2009numerically, Schubert2018numerically}.
We note however, that, as for SA, $a$ is typically much smaller than the total number of neurons, and the size of $n$ only matters at setup time, so for many real-world examples the latter optimizations are not really needed.

\paragraph{Our implementation of MDSA} We fit the covariance matrix using the estimator provided in \textsc{scikit-learn}, and its integrated function to calculate the Mahalanobis distance, which thus introduces only minimal technical debt in our code and leverages a well-tested, performance optimized implementation.

\subsubsection{Multi-Modal SA}
If the distribution of training set $AT_{SA}$ is multi-modal, i.e., can be interpreted as a set of different, \emph{simpler} (e.g. multivariate gaussian) distributions, is is reasonable to perform likelihood estimation taking advantage of multimodal distributions.
To do so, two modifications of SA have been proposed~\cite{Kim2021MultiModal}:
\emph{Multimodal LSA (MLSA)}, where instead of KDE, a \emph{Gaussian Mixture Model (GMM)} is used,
and \emph{Multimodal MDSA (MMDSA)}, where training set activation traces are clustered using the $k$-means algorithm, followed by the creation of a distinct instance of MDSA for each cluster.

\paragraph{Our implementation of Multi-Modal SA}
To implement MLSA, similar to LSA we use the \textsc{scikit-learn} implementation of GMM, which allows for a well-tested, fast and compact code.
To implement MMDSA, we introduce a general, abstract MultiModal-SA class architecture, based on the composite design pattern~\cite{Gamma1995Patterns}. While being a SA class instance itself, MultiModal-SA consists of a collection of sub-SA instances and a \emph{discriminator} to decide which sub-SA instance to use for a given input.
Fitting of both discriminator and sub-SA instances to the data is achieved by means of a transparent, single method call.
Given this abstract class, our MMDSA is then implemented by using the \textsc{scikit-learn} implementation of $k$-means as discriminator and our MDSA implementation for the sub-SA instances.
Provided that the different sub-SA instances are independent of each other and thus parallelizable, our implementation of the MultiModal-SA class allows the specification of the number of threads to be used when quantifying surprise.

\subsubsection{Per-Class SA}
Classification problems can be viewed as a special case of the MultiModal SA distributions described above.
Indeed, samples for which the DNN predicts the same class label form a cluster.
Thus, already when LSA was first  proposed, the authors recommended to use a dedicated instance of LSA for each class label~\cite{Kim2018}, intuitively leading to a better likelihood estimation.
We denote this approach as \emph{Per-Class LSA (PC-LSA)}.
Naturally, we would expect a similar advantage also for MDSA, MLSA and MMDSA, leading to \emph{PC-LSA}, \emph{PC-MDSA}, \emph{PC-MLSA} and \emph{PC-MMDSA}.
It is worthwhile that for SA variants for which quantification time grows in $n$, such as LSA, per-class quantification furthermore reduces quantification time, as only a subset of the training set $AT_{SA}$ has to be considered.

\paragraph{Our implementation of Per-Class SA} Given our  MultiModal-SA implementation described above, implementing per-class variants of SA is easily achieved by using the predicted class label as discriminator.

\subsubsection{Surprise Coverage}
To use surprise adequacy as a coverage criterion, the coverage profile is constructed as a one-dimensional boolean array, representing equally large, adjacent buckets of surprise. 
A bucket is covered by a test input if such input has a surprise adequacy in the range represented by the given bucket. 
Thus, for each considered DNN neuron the surprise coverage profile of an input can have either one or zero covered buckets, where the latter happens in  case the surprise value falls outside of the range of surprise values defined for the neuron's buckets. Once surprise coverage profiles are available, test prioritization can be achieved by applying the total or the additional method (CTM or CAM).

\subsection{DeepGini}
\label{sec:approach_deepgini}
Having massive scalability in mind, \emph{DeepGini}~\cite{Feng2020deepgini} provides a way to calculate a test prioritization score by working only on the test inputs activations of the DNNs softmax output layer (which also limits the applicability of DeepGini to classification problems).
Thus, DeepGini's runtime and memory requirements are only dependent on the number of classes, which  is typically small and which we consider a constant, making the runtime and memory requirments of DeepGini \BigO{1}.
Given a classification problem with $C$ classes, the softmax values $l_c(i)$ for class $c$ and input $i$, which are values between 0 and 1, summing up to 1, DeepGini is defined as
$$
\text{DeepGini}(i) = 1 - \sum_{c = 1}^C l_c(i)^2
$$
The DeepGini score is minimum (zero) when the DNN predicts one class with high certainty, assigning a softmax value of zero to all classes except the predicted one. It then increases as softmax values are distributed across an increasing number of classes that compete with the predicted one as classification alternatives.
When used as a TIP, DeepGini orders the  inputs by decreasing score, so as to give higher priority to  inputs with higher spread of softmax values across classes, as these inputs are associated with a higher classification uncertainty.

Besides its fast runtime, the design of DeepGini is primarily motivated by the property of having a single maximum, occurring when the predicted softmax value is the same for all classes, as proved by DeepGini's original authors~\cite{Feng2020deepgini}.

\subsection{Uncertainty Quantifiers}
\label{sec:approach_fault_pred}
As stated by its authors, ``DeepGini is designed [\ldots] to quickly identify misclassified tests''\cite{Feng2020deepgini}, i.e., to estimate the uncertainty of the DNN on the reported classification.
With a similar objective, the literature provides a range of DNN \emph{uncertainty quantifiers}~\cite{Weiss2021FailSafe}.
In this paper, we thus compare DeepGini also against such uncertainty metrics. 
Specifically, we consider \emph{(Softmax-)Entropy}, \emph{Vanilla-Softmax}, \emph{Prediction-Confidence Score} and \emph{Monte-Carlo Dropout}.

\paragraph{Vanilla Softmax} The Vanilla-Softmax metric is simply the highest activation in the output softmax layer for a classification problem,
subtracted from $1$ to obtain a metric that  correlates \emph{positively} with the misclassification probability, similar to DeepGini.
$$
\text{Vanilla Softmax}(i) = 1 - \max_{c=1}^C l_c(i)
$$

As the argmax of the softmax array is used as the DNNs predictions, Vanilla Softmax comes at virtually no computational or theoretical complexity and is thus often used as a naive, very simple benchmark~\cite{Weiss2021FailSafe, Ovadia2019, Berend2020cats, Hendrycks2016}.
Besides its simplicity, Vanilla Softmax also guarantees a single global extremum, as was aimed for when proposing DeepGini: 
Vanilla-Softmax reaches its single global maximum when all classes are predicted with the same softmax likelihood.
The proof sketch is simple:
Clearly, the hypothesized optimum ($l_c(i) = {}^1{\mskip -5mu/\mskip -3mu}_C,  \forall c \in \{1,...,C\}$) is a valid softmax array (i.e., all values are between $0$ and $1$, sum up to $1$ and the predicted class has the weakly highest value). 
Any  lower  value of the predicted class's softmax value, which is needed to further increase Vanilla-Softmax, would make it no more the highest, hence the predicted, value.
Similarly, to show that this is the \textit{unique} optimum we can observe that if the winning class has $l_c(i) = {}^1{\mskip -5mu/\mskip -3mu}_C$, all other classes must have the same value in a valid softmax array.

\paragraph{Prediction-Confidence Score (PCS)~\cite{Zhang2020}}
The \emph{Prediction-Confidence Score} is defined as the difference in softmax likelihood between the predicted class and the second runner-up class.
The main motivation behind PCS is that smaller values of PCS indicate that a prediction is made close to the decision boundary, i.e., a small change in the input might possibly  change the predicted class.
Same as for Vanilla-Softmax and DeepGini, when referring to PCS, we actually  subtract it from 1 to get a score which is expected to correlate \emph{positively} with misclassification probability~\cite{Weiss2021FailSafe}.

\paragraph{Entropy} 
The authors of DeepGini also considered the entropy in the softmax layer as an alternative to their approach~\cite{Feng2020deepgini}. 
No empirical comparison between the two was made however, as the authors motivated their preference for DeepGini claiming that it is a simpler metric, better justified for DNN test prioritization and requiring no information theory background and a `non-statistical view'~\cite{Feng2020deepgini} to the problem.

\paragraph{MC-Dropout}
Point-predictor based uncertainty metrics, such as Vanilla-Softmax, PCS and Entropy, while showing very good misclassification prediction results in practice~\cite{Weiss2021FailSafe}, suffer from major theoretical drawbacks when used as uncertainty quantifier~\cite{Gal2016thesis}.
A simple, yet theoretically much better founded way to extract uncertainty is \emph{Monte-Carlo Dropout}, where dropout layers used for stochastic regularization during training are enabled at prediction time, which allows the sampling of multiple stochastic predictions for the same test input and correspondingly for the inference of both an average prediction and the associated uncertainty (e.g., standard deviation) from the observed distribution of predicted values~\cite{Gal2016dropout}. 

\paragraph{Our implementation of uncertainty quantifiers} We base our implementation of uncertainty metrics on \textsc{uncertainty-wizard}~\cite{Weiss2021uwiz}, which provides configurable, tested and fast implementations of various uncertainty metrics.
It also exposes interfaces to add new uncertainty metrics, which we used to implement DeepGini.

\section{Empirical Procedure}
\label{sec:emp_proc}
The \textit{goal} of our experiments is to compare the previously described TIP approaches in various configurations and along three dimensions: (1) runtime, (2) test prioritization effectiveness, and (3) active learning capabilities.
The aim is to replicate the original DeepGini results, but also to extend them to uncertainty quantifiers, not considered in the original DeepGini paper~\cite{Feng2020deepgini}, and to more recent variants of SA. We also make sure that the non determinism affecting the training process is taken into account by statistical tests, not carried out in the original DeepGini paper.
Correspondingly, we answer the following research questions:
\begin{description}[noitemsep]
\item [RQ\textsubscript{1}: Replication] Are our results consistent with those obtained by the authors who proposed DeepGini, i.e., does DeepGini outperform NC, LSC and DSC, along the three considered dimensions?
\item [RQ\textsubscript{2}: Comparison to Other Approaches] How does DeepGini compare to uncertainty quantifiers and newer SA variants?
\item [RQ\textsubscript{3}: Statistical Analysis] Are our findings sensitive to random influences due to the DNN training process?
\end{description}

\subsection{Experimental Design}
As we are replicating an existing paper, the broad context of our analysis is predetermined by the original paper.
Within such constraints however, we made the following modifications to mitigate some of the original threats to validity, leading to overall more reliable results.

\begin{itemize}[noitemsep,leftmargin=*]
    \item \emph{Reimplementation:} We implemented the scripts to run our experiments without consulting or copying from the reproduction package provided alongside the replicated paper~\cite{Feng2020deepgini}. This reduces the risk of having the same bugs in our code and in the original code, and follows the ACM guidelines for reproducibility~\cite{ACMReproducibility}.
    \item \emph{Repeated Runs:} Random influences in the DNN training can have major influences in the performance measured by adequacy criteria~\cite{Weiss2021-SA}. We thus repeat all our experiments 100 times and discuss statistical significance in RQ\textsubscript{3}.
    \item \emph{Additional Case Study and Reimplemented Model Architectures:} Replicability studies are also supposed to check that under reasonable changes that extend the validity scope, the overall findings remain unchanged~\cite{ReScienceCReproducibility}. We thus replaced one case study (Svhn) with a different one (Imdb) and implemented our own versions of the DNN model architectures for the case studies, to see if the replicated findings indeed generalize to a new case study and to different implementations of the original case studies.
    \item \emph{Original threats to validity:} We identified  issues (e.g., with adversarial examples) in the original experimental design, which may give DeepGini an unfair advantage. We thus changed the experimental design to fix them (see below).
\end{itemize}

\subsubsection{Timing Experiments}
Timing experiments were conducted on the first 10 (of the 100)  test prioritization experiments run for the Mnist case study. 
While timing was taken, the machine was otherwise not used, to ensure an as fair as possible comparison between the different approaches.
For the remaining experiments, times were also tracked, and are available in our reproduction package, but for reasons of efficiency, we ran multiple runs in parallel, which  makes these runtime measures not comparable.

\subsubsection{Test Prioritization}
Our experimental design follows the design described in the replicated paper: 
Test datasets are prioritized by all compared approaches and are evaluated for their ability to give high priority to examples that make the DNN fail (in the following, we equate a DNN failure with a DNN misclassification, as DeepGini is applicable only to classifier DNNs). 
In testing terminology, each misclassified input would be considered a (unique) fault, and the approach which detects most faults as early as possible is preferable. 
This is measured using a  standard metric in test prioritization, the \emph{Average Percentage Faults Detected} (APFD)~\cite{ElbaumMR00,LiangER18,RothermelUCH01}.
Every approach is tested on a nominal test set, which contains just a few misclassifications, and on an out-of-distribution (OOD) test set, containing additional misclassifications.
The OOD test set consists of the nominal  inputs, extended with additional, harder to classify inputs. For the latter, the original paper uses adversarial examples, while we use corrupted inputs, as motivated below.

\paragraph{Problem with Adversarial Examples} The replicated paper uses adversarial examples to create OOD, misclassified data points. 
While using adversarial examples has some  advantages, amongst which are a very easy way to create arbitrary input formats using tools like \emph{foolbox}~\cite{rauber2017foolbox, rauber2017foolboxnative} and their guaranteed misclassification, we argue that they come with a major threat to validity when comparing DeepGini against Neuron Coverage and Surprise Adequacy:
Most adversarial attacks perturbate the input in a minimal way, which is still sufficient to produce the desired model output (typically a misclassification). 
DeepGini, but also Max-Softmax, Entropy and PCS, compute their priority score based exactly on the output that is deliberately changed by an adversarial attack. Since adversarial attacks perform minimal changes necessary to achieve a misclassification, typically the output softmax distribution under an adversarial attack is reshaped until one of the wrong classes becomes the winner, leaving the other classes with a relatively high softmax value, which makes the input easily detectable by DeepGini, Max-Softmax, Entropy and PCS, giving them an unfair advantage over the other approaches.
Moreover, the adversarial input perturbations introduced by adversarial attacks cannot be considered as representative of real world test inputs, as they are synthesized artificially based on the internal DNN operations to fool the final classification.
Hence, with adversarial attacks as OOD test set it would be unclear if the observed results generalize to natural, more realistic OOD inputs.

\paragraph{Corrupted Data} 
To mitigate the threat to validity which comes with adversarial data, we use \emph{corrupted inputs} instead:
Corrupted inputs are test sets where nominal data points are manipulated using a range of modifiers  inspired by real-world input corruptions.
They thus provide DNN-independent, realistic OOD data. 
While the literature uses and provides a range of corrupted test sets~\cite{Hendrycks2018, Mu2019, Stocco2020}, for two of our case studies no corrupted datasets existed yet and we thus propose novel corrupted datasets for them. These are described in \autoref{sec:test_subjects}.

\subsubsection{Active Learning}

\begin{figure}
    \centering
    \includegraphics[width=0.8\linewidth]{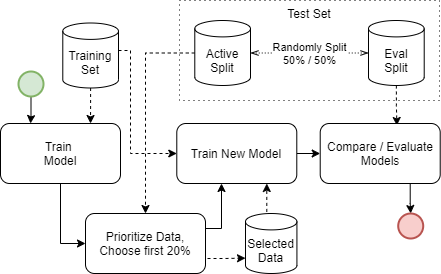}
    \caption{Active Learning Experimental Setup}
    \label{fig:exp_des_act_learning}
\end{figure}

As in the replicated paper, we split our test sets into two parts, the \emph{active} split, from which 20\% of the nominal dataset samples (or 10\% of the OOD dataset samples, to obtain the same absolute number of selected samples)
are used for retraining, and the \emph{eval} split, which is used for final evaluation of the retrained models. This process is illustrated in \autoref{fig:exp_des_act_learning}.
As opposed to the original authors, we perform the active learning experiments not only using OOD test sets, but also using nominal test sets.
Moreover, we also measure the cross-dataset improvements, i.e., how well a nominal active split helps to improve performance on an OOD eval split, and vice-versa.
Lastly, our experiments further differ from the ones conducted in the replicated in that we compare the TIP-based training set selection against a random training set selection, to measure the actual benefit gained by using a TIP.

\subsection{Tested Approaches}
We compare DeepGini against a total of 38 TIP, consisting of 24 different NC TIP, 10 surprise TIP and 4 uncertainty TIP.
For what concerns the NC TIP, we base our selection of $k$ on the choices made in the replicated paper, with the exception of KMNC:
Here, the authors used $k=1,000$ and $k=10,000$, but these values lead to coverage profiles so large that the approaches did not terminate in a reasonable amount of time in our setting, where multiple runs are carried out to gain statistical significance.
To be able to include it, we used KMNC with $k=2$, leading to a size of the coverage profiles comparable to that obtained with the other NC metrics.
We test all our NC metrics using both CAM and CTM.

For what concerns SA, we chose the same number of segments in the coverage profiles as the replicated paper (1,000), but we dynamically select the upper bound used to define these segments based on the surprise adequacy values observed on the test dataset, as we assume would be done in a practical setting.
As thus every test input has the exact same coverage ($1/1000$), using CTM would be equivalent to random ordering.
For SA, we thus consider CAM on the surprise coverage profiles, but also the raw surprise adequacies in decreasing order as a replacement for CTM, such that the most surprising inputs would be selected first.
As DeepGini is only applicable to classification problems, we use the per-class variants of LSA, MLSA, MDSA and MMDSA. 

For what concerns  the uncertainty metrics, only MC-Dropout has configurable parameters, and we use a very large number of samples (200), aggregated using variation ratio, as recommended in recent literature~\cite{Weiss2021FailSafe}.
As opposed to the replicated paper, which further narrowed down their selection of tested approaches after collecting initial insights, we evaluate all TIP approaches for all research questions.

\begin{table*}[t]
    \centering
    \small
    
\newcommand{\best}{\cellcolor{lightgray!40}}
\newcommand{\verti}[1]{\begin{tabular}{@{}c@{}}\rotatebox[origin=c]{90}{\centering #1}\end{tabular}}\begin{tabular}{llcccccccccccc}
\toprule
            &            & \multicolumn{3}{c}{mnist} & \multicolumn{3}{c}{fmnist} & \multicolumn{3}{c}{cifar10} & \multicolumn{3}{c}{imdb} \\
            &            & nominal &    ood &   time & nominal &    ood &   time & nominal &    ood &   time & nominal &    ood &  time \\
 & approach &         &        &        &         &        &        &         &        &        &         &        &       \\
\midrule
\multirow{8}{*}{\verti{neuron coverage}} & NAC-0.75-CAM &  43.83\% & 48.71\% &    17s &  42.46\% & 45.80\% &    24s &  48.88\% & 46.70\% &    72s &  26.55\% & 30.82\% &   20s \\
            & NAC-0.75 &  38.64\% & 44.97\% &    11s &  41.25\% & 45.35\% &    19s &  48.58\% & 46.52\% &    62s &  26.53\% & 30.82\% &   19s \\
            & NBC-0-CAM &  56.89\% & 53.33\% &    53s &  42.91\% & 48.79\% &    80s &  49.59\% & 50.78\% &   200s &  60.08\% & 54.01\% &   34s \\
            & NBC-0 &  54.03\% & 44.50\% &    43s &  38.69\% & 45.48\% &    68s &  49.18\% & 50.80\% &   183s &  60.09\% & 54.02\% &   34s \\
            & SNAC-0-CAM &  51.61\% & 69.46\% &    43s &  50.37\% & 59.45\% &    70s &  50.36\% & 49.97\% &   179s &  51.33\% & 50.01\% &   34s \\
            & SNAC-0 &  51.30\% & 70.36\% &    37s &  49.71\% & 59.63\% &    63s &  50.31\% & 49.88\% &   174s &  51.32\% & 50.01\% &   34s \\
            & TKNC-1-CAM &  49.75\% & 56.02\% &    42s &  50.48\% & 53.74\% &    58s &  50.13\% & 50.61\% &   109s &  51.43\% & 50.04\% &   20s \\
            & KMNC-2 &  40.96\% & 29.76\% &    39s &  49.52\% & 40.13\% &    64s &  50.17\% & 50.34\% &   177s &  51.41\% & 50.05\% &   33s \\
\cline{1-14}
\multirow{5}{*}{\verti{surprise}} & DSA &  92.48\% & 69.73\% &  1060s &  76.71\% & 71.48\% &  1012s &  61.17\% & 57.93\% &  1143s &  66.20\% & 65.18\% &   43s \\
            & PC-LSA &  87.47\% & 65.13\% &   152s &  60.99\% & 64.26\% &   148s &  49.77\% & 47.28\% &   135s &  55.83\% & 56.14\% &   29s \\
            & PC-MDSA &  89.86\% & 68.64\% &    60s &  64.34\% & 64.29\% &    59s &  56.13\% & 52.98\% &   128s &  65.78\% & 64.14\% &    \best3s \\
            & PC-MLSA &  89.59\% & 69.16\% &   107s &  64.06\% & 63.08\% &    93s &  52.69\% & 50.73\% &   107s &  55.95\% & 57.60\% &    5s \\
            & PC-MMDSA &  79.94\% & 66.28\% &   134s &  56.23\% & 63.90\% &   121s &  51.39\% & 48.90\% &   191s &  60.43\% & 60.80\% &   12s \\
\cline{1-14}
\multirow{5}{*}{\verti{uncertainty}} & DeepGini & \best98.22\% & 87.80\% &     \best1s &  86.35\% & 72.36\% &    \best 2s &  70.89\% & \best 68.90\% &     \best3s &  \best73.52\% &\best 68.93\% &   \best 3s \\
            & Vanilla SM &  \best 98.22\% & 88.09\% &     \best1s &  \best86.42\% & 72.26\% &    \best 2s &  70.81\% & 68.79\% &    \best 3s & \best 73.52\% &\best 68.93\% &    \best3s \\
            & PCS &  \best 98.21\% & 88.61\% &    \best 1s &  86.28\% & 71.80\% &     \best 2s &  70.55\% & 68.47\% &     \best3s &  \best73.52\% & \best68.93\% &    \best3s \\
            & Entropy &  \best 98.20\% & 87.27\% &    \best 1s &  86.03\% & 72.47\% &    \best 2s &  \best 70.97\% & 69.03\% &     \best3s &  \best73.52\% &\best 68.93\% &   \best 3s \\
            & MC-Dropout &  97.35\% & \best90.59\% &   221s &  83.71\% & \best73.07\% &   392s &    n.a. &   n.a. &   n.a. &  60.38\% & 56.56\% &  364s \\

        \bottomrule
        \multicolumn{14}{c}{\begin{tabular}{@{}c@{}}
        This table provides an overview of selected results.
        Find the full table, including 21 additional TIP, as CSV, in the \replipkg. \\
        APFDs at most $0.05\%$ worse than the best value, or times at most 1s higher than the fasted are highlighted gray.
        \end{tabular}}\\
        \bottomrule
        
\end{tabular}

    \caption{Test Prioritization APFDs (capability to detect misclassifications) and Runtimes}
    \label{tab:apfd}
\end{table*}

\subsection{Test Subjects}
\label{sec:test_subjects}

\paragraph{Mnist} 
A grey-scale image digit classification problem~\cite{lecun2010mnist}, representing the most commonly used dataset in the literature on testing of machine learning based systems~\cite{Riccio2020}. As corrupted data, we use Mnist-c~\cite{Mu2019}.
\paragraph{Fashion Mnist (Fmnist)} 
A more challenging drop-in replacement for Mnist, representing 10 different types of clothing~\cite{Xiao2017fmnist}. To the best of our knowledge, no corrupted version exists in the literature. Thus, we created and plan to release (in case of paper acceptance) \emph{Fashion-mnist-c}, where images are corrupted using selected corruption methods from Mnist-c, as well as  additional corruptions targeting image orientation and rotations, since most nominal images are identically oriented. 
Examples of corruptions are: Various types of noise (e.g. shot-noise), blurring (e.g. glass-blurring), transformations (e.g. saturation, brightness), and orientation (e.g. flipping right-left).

\paragraph{Cifar-10} A ten class color image classification problem~\cite{Krizhevsky09cifar10}. As corrupted data, we use Cifar10-c~\cite{Hendrycks2018}.

\paragraph{Imdb} Binary sentiment (positive / negative) classification of textual Imdb reviews. 
For classification, we use a transformer model, thus an architecture very different from the one used in the other case studies.
For the Imdb dataset to the best of our knowledge no corrupted datasets are available from the literature. We thus created and plan to release (in case of paper acceptance) a text corruption approach, which corrupts text by mimicking wrong auto-completions (replacing words with other words starting with the same 3 letters), wrong auto-corrections (replacing words with other words at small Levenshtein distance~\cite{levenshtein1966binary}),  bad single-word level translations (replacing words with a synonym based on Wordnet~\cite{fellbaum2010wordnet}, potentially ill-chosen given the context), and single-letter typos.
Our approach is configurable by severity level, to generate instances of \emph{Imdb-c} with corruptions of variable severity.

\section{Results}
\label{sec:results}

\subsection{RQ\textsubscript{1} Replication}

\subsubsection{Execution Time}
The fact that DeepGini is much faster to compute than both NC and DSC, LSC  is obvious from the approach descriptions in \autoref{sec:approaches}, 
and our empirical results confirmed this fact.

The results (time required to prioritize a full test set, averages over the nominal and ood test sets) are shown in \autoref{tab:apfd}.
Prioritizing a mnist test set took less than 2 seconds on average for DeepGini (which essentially reflects the time required for the DNN to make the softmax predictions). 

In the same setting, the NC-CTM methods required between 11 and 39 seconds, with an additional 2 to 12 seconds if CAM was used.
DSA and PC-LSA took much longer than even NC: 
For example, for mnist, PC-LSA took 152s, DSA took more than 17 minutes (1060s). 
Clearly, PC-MDSA is the fastest SA metric, taking only 60s to prioritize the same dataset.
Given the simple nature of the surprise coverage profiles (only 1,000 booleans, out of which exactly one is \textsc{true}), the calculation of the surprise coverage CAM variants took less than a second longer than their surprise adequacy counterparts (a table with the detailed results is omitted for space reasons, but is included in the replication package).
Understandably, our absolute measurements differ from the ones reported in the replicated study~\cite{Feng2020deepgini}, amongst other reasons as we use different implementations, different hardware, different OOD test sets and different models. 
However, for what concerns their primary finding, i.e.,  DeepGini being the fastest approach, we  come to the same conclusion.

\subsubsection{Test Prioritization}
The results for the test prioritization experiments are reported in \autoref{tab:apfd}. 
The approaches which were originally compared against DeepGini, i.e., the NC variants, as well as the DSC and PC-LSC  variants of surprise (not reported in \autoref{tab:apfd} for lack of space, but just mildly inferior to DSA and PC-LSC, respectively), perform worse than DeepGini on all case studies on both nominal and OOD test datasets, confirming the primary finding in the replicated paper~\cite{Feng2020deepgini}:
First, with the expected APFD value for random ordering being 50\% (asymptotically, for a large number of prioritized tests), we can see that DeepGini is clearly effective at identifying misclassified DNN inputs, as it's lowest observed APFD value is 68.9\%.
The NC metrics, in most cases, do not perform much better, if better at all, than the random baseline in both their CTM and CAM variants. 
Similarly, the CAM variants of the basic surpise adequacies, PC-LSA-CAM and DSA-CAM, performed worse than DeepGini, but better than most NC TIP, which is in line with the findings in the replicated paper.

Here, it is worth mentioning that while the original experiments~\cite{Feng2020deepgini} only measured the APFDs on the Mnist case study, 
their primary finding that DeepGini outperforms all other approaches w.r.t. APFD values generalizes to all our case studies, as well as to the corruption based OOD datasets.
For what concerns the latter, as expected we find that  adversarial examples used in the original experiments~\cite{Feng2020deepgini} are much more favorable towards DeepGini than the more realistic corruption based outliers that we used in our study: 
The performance of DSA-CAM and (PC-)LSA-CAM on the nominal Mnist test set is similar in both papers with respect to their absolute APFD value (roughly 65\%). However, DeepGini achieved an APFD of >98\% in the experiments with adversarial data~\cite{Feng2020deepgini}, while using the corruption based OOD data, the APFD performance of DeepGini dropped by more than 10\%.

\subsubsection{Active Learning}

The results for the active learning experiments are reported in \autoref{tab:active}. 
Differently from \replicatedPaper, we report the difference in accuracy to a random data selection (RANDOM), to not only investigate which TIP performs best, but also  if and how much each TIP is effective in guiding active learning.
For comparison, the row "ORIGINAL" refers to the model accuracy before active learning.
For what concerns the dominance of DeepGini to guide active learning, our results are less clear than the ones reported by \replicatedPaper: 
For  Mnist, Fmnist and Cifar10, DeepGini outperforms all approaches considered by \replicatedPaper with only very view exceptions, but sometimes by a very narrow margin.
For Imdb, in particular when facing corrupted (OOD) inputs, we find that DeepGini is clearly outperformed by DSA and PC-MDSA. 
This may be caused by the different nature of this case study (text instead of image classification) or the correspondingly different type of OOD data.

In all case studies,  we found that  NC techniques are sometimes inferior to the random baseline.
This showcases the importance to compare each approach with a random baseline. 
Without it, all TIP would have appeared successful at guiding active learning.

Clearly, our results are more ambiguous than the ones reported in \replicatedPaper.
A primary root cause for this is  our  empirical setup, representing a more realistic setting with less and diverse outliers, which pose a much harder, but more realistic problem for active learning.
However, we still clearly observe DeepGini's capability to prioritize inputs for active learning, and doing so better than neuron and surprise coverage in most cases, and we thus still consider our results a successful confirmation of the superiority of DeepGini reported in \replicatedPaper.

\subsubsection{Confirmed findings from other papers}
While not the primary objective of our experiments, our results confirm various findings reported in recent research papers, in particular w.r.t. surprise adequacy:
First, we find that MDSA is indeed a valuable, faster alternative to LSA, yielding comparable if not better results in our experiments, which is in line with the results in the paper proposing MDSA~\cite{Kim2020MDSA}.
Second, the fact that DSA typically outperforms PC-LSA is consistent with the paper proposing these two approaches~\cite{Kim2018}.
Lastly, we also observed the previously reported non-dominance between PCS and MC-Dropout~\cite{Zhang2020}.
    
\begin{tcolorbox}[colback=cyan!5,colframe=cyan!75!black,title=Summary of RQ\textsubscript{1} (Replication)]
Our experimental results fully support the dominance of DeepGini on execution times and test prioritization effectiveness, and with only a few justified exceptions, active learning capability. 
In addition, some of our results are consistent with a range of observations made in related papers~\cite{Kim2018, Kim2020MDSA, Zhang2020}.
\end{tcolorbox}

\begin{table*}[]
    \centering
    \small
    \newcommand{\best}{\cellcolor{lightgray!40}}
\newcommand{\verti}[1]{\begin{tabular}{@{}c@{}}\rotatebox[origin=c]{90}{\centering #1}\end{tabular}}\begin{tabular}{llcccccccc}
\toprule
            &            & \multicolumn{2}{c}{mnist} & \multicolumn{2}{c}{fmnist} & \multicolumn{2}{c}{cifar10} & \multicolumn{2}{c}{imdb} \\
            &            &        nominal &        ood &        nominal &        ood &        nominal &        ood &        nominal &        ood \\
 & approach &                &            &                &            &                &            &                &            \\
\midrule
\multirow{2}{*}{\verti{baseline}} & ORIGINAL &         99.15\% &     93.68\% &         90.15\% &     72.17\% &         69.16\% &     64.74\% &         83.72\% &     80.28\% \\
            & RANDOM &         99.15\% &     95.52\% &         90.12\% &     75.20\% &         69.42\% &     65.66\% &         83.95\% &     81.45\% \\

                          &&\multicolumn{8}{c}{\footnotesize{Subsequent results are differences to the \emph{random} baseline.}}\\
                          \cline{1-10}
                          
\multirow{8}{*}{\verti{neuron coverage}} & NAC-0.75-CAM &          0.01\% &      0.03\% &          0.08\% &      0.33\% &          \best0.14\% &      0.12\% &         -0.39\% &     -0.64\% \\
            & NAC-0.75 &         -0.00\% &     -0.86\% &          0.05\% &     -0.62\% &          0.12\% &     -0.03\% &         -0.26\% &     -0.78\% \\
            & NBC-0-CAM &          0.01\% &      0.26\% &          0.09\% &      0.71\% &         -0.02\% &      0.09\% &          0.14\% &      0.31\% \\
            & NBC-0 &          0.01\% &     -1.05\% &          0.03\% &     -0.93\% &         -0.23\% &     -0.15\% &          0.15\% &      0.28\% \\
            & SNAC-0-CAM &          0.02\% &      0.09\% &          0.16\% &      0.75\% &         -0.02\% &     -0.01\% &          0.02\% &      0.01\% \\
            & SNAC-0 &          0.02\% &      0.14\% &          0.11\% &      0.58\% &          0.09\% &      0.15\% &         -0.06\% &      0.02\% \\
            & TKNC-1-CAM &          0.02\% &      0.32\% &          0.09\% &      0.82\% &         -0.10\% &      0.08\% &         -0.03\% &      0.07\% \\
            & KMNC-2 &          0.00\% &     -1.87\% &          0.08\% &     -1.96\% &         -0.01\% &     -0.02\% &          0.04\% &     -0.02\% \\
\cline{1-10}
\multirow{5}{*}{\verti{surprise}} & DSA &          0.04\% &      0.80\% &          0.11\% &      1.41\% &          0.05\% &      0.38\% &          0.24\% &      \best0.48\% \\
            & PC-LSA &          0.01\% &      0.11\% &          0.07\% &      0.78\% &         -0.10\% &      0.23\% &          0.23\% &      0.20\% \\
            & PC-MDSA &          0.03\% &      0.04\% &         \best 0.14\% &      0.64\% &         -0.05\% &      0.36\% &          0.24\% &      \best 0.47\% \\
            & PC-MLSA &          0.03\% &      0.17\% &          0.13\% &      0.62\% &         -0.06\% &      0.12\% &          0.21\% &      0.32\% \\
            & PC-MMDSA &          0.02\% &      0.17\% &          0.12\% &      0.74\% &          0.05\% &      0.10\% &          0.18\% &      0.37\% \\
\cline{1-10}
\multirow{5}{*}{\verti{uncertainty}} & DeepGini &          \best0.07\% &      0.89\% &          0.12\% &      1.30\% &          0.10\% &      0.57\% &          0.17\% &      0.37\% \\
            & Vanilla SM &          0.04\% &      0.90\% &          0.13\% &      1.44\% &          0.04\% &      0.42\% &          \best0.29\% &      0.38\% \\
            & PCS &          0.05\% &      1.02\% &          \best 0.14\% &      1.16\% &          0.06\% &      0.40\% &          0.22\% &      0.31\% \\
            & Entropy &          0.05\% &      0.82\% &          0.11\% &      1.37\% &         -0.06\% &      \best 0.55\% &          0.18\% &      0.38\% \\
            & MC-Dropout &          0.05\% &     \best 1.22\% &         \best 0.15\% &      \best 1.53\% &           n.a. &       n.a. &          0.20\% &      0.35\% \\

        \bottomrule
        \multicolumn{10}{c}{\begin{tabular}{@{}c@{}}
         This table provides an overview of selected results. Find the full table, \\
        including 21 additional TIP, cross dataset and cross-split evaluations in the \replipkg.\\
        Accuracies at most 0.01\% worse than the best value are highlighted gray.
        \end{tabular}}\\
        \bottomrule
        
\end{tabular}

    \captionof{table}{Active Learning (capability to prioritize data for retraining)}
    \label{tab:active}
\end{table*}

\subsection{RQ\textsubscript{2} Comparison}
Overall, we considered three additional types of TIP in our study, which were not considered in the study by \replicatedPaper:
(1) The use of  \emph{raw} surprise adequacy instead of  surprise coverage, (2) Recently proposed variants of surprise adequacy, specifically PC-MDSA, PC-MLSA, PC-MMDSA, and (3) uncertainty quantifiers.

For what concerns surprise adequacy, our results (see \autoref{tab:apfd} and \autoref{tab:active}) show that using  raw surprise adequacy instead of surprise coverage is beneficial, which is expected: 
Test inputs with high surprise adequacy are more novel and unexpected, and thus intuitively also more challenging to classify and more useful for active learning.
Furthermore, we find that as intended when proposed, MDSA offers a faster replacement of LSA with similar prediction performance, and that multimodal SA can indeed improve performance over unimodal SA, as we can see when comparing MLSA with LSA in \autoref{tab:apfd} and \autoref{tab:active}. 
Despite all improvements, DSA remains the best performing SA variant, while all SA variants are in general inferior to DeepGini.

On the other hand, we find that DeepGini is frequently, but not always, outperformed by the other uncertainty metrics we added to this study (see \autoref{tab:apfd} and \autoref{tab:active}).
While the fact that a sophisticated, theoretically well founded and very frequently used MC-Dropout performs well is not so surprising, it is quite unexpected that even the most simple, naive baseline, Vanilla Softmax, reaches similar performance as DeepGini.
Indeed, as discussed in RQ\textsubscript{3}, within the family of softmax-based uncertainty metrics, including DeepGini, we do not find statistically significant differences among the considered approaches.

\begin{tcolorbox}[colback=cyan!5,colframe=cyan!75!black,title=Summary of RQ\textsubscript{2} (Comparison to Other Approaches)]
We found that raw surprise adequacy often outperforms surprise coverage and that recently proposed innovations such as Multi-Modal or Mahalanobis Distance based SA are valuable alternatives to traditional LSA. 
Most interestingly, however, we found that DeepGini does not in general perform better than even the simplest baseline from uncertainty quantification, Vanilla Softmax. Actually, it is often outperformed by  uncertainty quantification approaches.
\end{tcolorbox}

\captionsetup{justification=centering}
\begin{figure*}
  \begin{minipage}{.45\textwidth}
    \includegraphics[width=\textwidth]{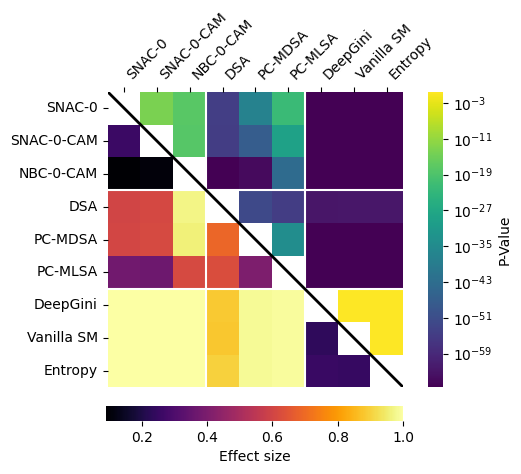}
    \captionof{figure}{Statistical Analysis: Test Prioritization}
    \label{fig:stats_heatmap}
  \end{minipage}
    \hfill
  \begin{minipage}{.45\textwidth}
    \includegraphics[width=\textwidth]{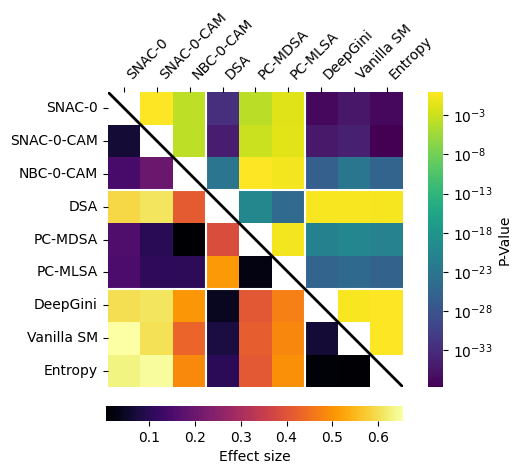}
    \captionof{figure}{Statistical Analysis: Active Learning}
    \label{fig:stats_active_heatmap}
  \end{minipage} \quad
\end{figure*}

\subsection{RQ\textsubscript{3} Statistical Analysis}
We pairwise compare all 39 tested TIP to assess the significance (using two-sided Wilcoxon signed-rank test) and effect size (using the paired Vargha-Delaney method) in their difference.
We applied Bonferroni error correction to account for the multiple tests\footnote{To allow easy comprehension of our results, we multiplied all p-values with the Bonferroni correction factor: ${39 \choose 2} = 741$.}.

\paragraph{Test Prioritization} Our statistical analysis regarding test prioritization's APFDs for the three best uncertainty quantification, surprise adequacy and neuron coverage approaches is shown in \autoref{fig:stats_heatmap} (the values for all approaches will be available in the replication package).
For what concerns the APFD values, our main findings from RQ\textsubscript{1} and  RQ\textsubscript{2} are confirmed:
The softmax based metrics DeepGini, Vanilla Softmax and Softmax Entropy outperform the other approaches with very low $p$-value and very high effect size.
Moreover, the low effect size and high $p$-value observed within this family of approaches (lower right 3x3 corner in \autoref{fig:stats_heatmap}) strongly supports our finding that DeepGini is not in general capable of outperforming the simplest  techniques from uncertainty quantification.

\paragraph{Active Learning} 
When running a similar analysis for our active learning experiments, we observe a similar heatmap, shown in \autoref{fig:stats_active_heatmap} (note the different ranges in the two axes w.r.t. \autoref{fig:stats_heatmap}).
Again, significance and effect size of  the differences within the uncertainty quantification family, which includes DeepGini, MC-Dropout and Softmax Entropy, is low, but in this case we find that this group extends to DSA, whose differences with respect to the simple uncertainty quantifiers are also insignificant and of low effect size. 

Active learning results have generally low effect sizes, indicating that the performance of most approaches is very similar.
This might be due to the challenging, but more realistic, setting adopted in our experimental design, where outliers are less frequent and more realistic than the large amount of adversarial examples used in the replicated paper.
In our setting, active learning   leads  to just marginal improvements and random influences during training largely impact the final performance.

\begin{tcolorbox}[colback=cyan!5,colframe=cyan!75!black,title=Summary of RQ\textsubscript{3} (Statistical Analysis)]
We can confirm the replicated paper's main finding, i.e., DeepGini's dominance over SA and NC, with high significance and large effect size.
However, there's no significant difference and negligible effect size when comparing DeepGini to simple baselines, such as Vanilla Softmax.
\end{tcolorbox}

\section{Threats to Validity}
\label{sec:threads}

\paragraph{Internal validity} 
When configuring the approaches being compared, we made several choices on their parameters. Whenever possible we remained consistent with the choices made in the paper being replicated~\cite{Feng2020deepgini}.
In other cases, we chose hyperparameters compatible with the time and resources allocated to our empirical comparison. E.g. KMNC-1000 and KMNC-10000 which were found to be too computationally expensive by \replicatedPaper, were replaced with KMNC-2, in order to achieve similar coverage profile size, and thus runtime, as most other NC approaches.
In the other cases, we document carefully and motivate our decisions in the paper.

\paragraph{Conclusion validity} 
Differently from the paper being replicated~\cite{Feng2020deepgini}, we mitigate the conclusion validity threat associated with the non-determinism of the training process by repeating our experiments 100 times, each time with a newly trained model. Thanks to such repetitions, our findings are corroborated by a statistical analysis, which discriminates differences due to chance from differences that are statistically significant.

\paragraph{External validity} 
Our results are focused on the problem of test input prioritization for early fault detection or active learning, so they might not generalize to other problems that require test input ordering or selection.
In particular, while our experiments suggest the use of simple softmax-based TIP, such as Vanilla Softmax, PCS, Softmax-Entropy, but also DeepGini,
they should not be blindly trusted for arbitrary applications that deviate from the ones considered in this paper.
In fact, when used as uncertainty metric, Vanilla Softmax suffers from  well known theoretical drawbacks~\cite{Hendrycks2016, Gal2016thesis, nguyen2015deep}.
Much of this critique revolves around Vanilla Softmax's inability to reliably quantify the absolute probability of a fault (values are typically underestimated) and it being a point-predictor (i.e., non-Bayesian) approach. It is also easily fooled by adversarial examples.

By replacing Svhn with Imdb as our fourth case study, we have extended the external validity of the findings reported in the original paper~\cite{Feng2020deepgini} to a quite different domain, natural language processing for sentiment analysis. Still, the overall number of case studies across both papers remains limited, which demands
for additional, similar work on different case studies which we support by providing a comprehensive \replipkg.

\section{Artifacts}
\label{sec:artifacts}

Alongside our paper, we release a set of artifacts in a way which facilitates not only reproduction of our results, but also  easy and reliable reuse of our code possibly for different purposes.
To this extent, we release four kinds of artifacts: 
(1) a full reproduction package, (2) software libraries that we developed for core  components of the evaluated approaches, (3) datasets, and (4) all experimental results as CSV tables.
All our artifacts are released under a permissive MIT license and archived on \href{https://zenodo.org/}{zenodo.org}.

\paragraph{Reproduction Package}
All our code is available on a Github repository and through a pre-built Docker image. 
The latter exposes a command line interface to easily re-run all or selected parts of our experiments, without having to install all dependencies manually.
Alongside the full code, we release an archive of files containing intermediate results of our experiments, such as trained models and calculated priorities.

\begin{center}
    \textbf{Image:}\small{ \texttt{docker pull ghcr.io/testingautomated-usi/simple-tip}}\newline
    \textbf{Code \& Docs:} \href{https://github.com/testingautomated-usi/simple-tip}{github.com/testingautomated-usi/simple-tip}.
\end{center}

\paragraph{\textsc{dnn-tip}: DNN Test Prioritizer Software Library}
We release core components of our experiments, allowing to use and assess different NC and SA metrics. These are  
likely to be useful also in other, different studies (e.g., our implementations of the different NC and SA approaches), being provided
as an easy to use, standalone Python library.
\begin{center}
    \textbf{Install from pypi}: \texttt{pip install dnn-tip} \newline
    \textbf{Code \& Docs:} \href{https://github.com/testingautomated-usi/dnn-tip}{github.com/testingautomated-usi/dnn-tip}.
\end{center}
We merged our implementation of DeepGini, which can be seen as a variant of common uncertainty quantifiers, into \textsc{uncertainty-wizard}, a library providing a wide range of uncertainty quantifiers~\cite{Weiss2021uwiz, Weiss2021FailSafe}.

\paragraph{Datasets}
Corrupted datasets, such as \textsc{Mnist-c}~\cite{Mu2019}, \textsc{Cifar-10-c}~\cite{Hendrycks2018} and \textsc{Imagenet-C}~\cite{Hendrycks2018} have been shown to be useful in a wide range of studies regarding robustness of DNNs, and we thus release our \textsc{Fmnist-c} and \textsc{Imdb-c} datasets as standalone artifacts:
\textsc{Fmnist-c} can easily be installed through Huggingface datasets~\cite{Lhoest2021HuggingfaceDS}:
\begin{center}
    \textbf{Huggingface}: \texttt{mweiss/fashion\_mnist\_corrupted} \newline
    \textbf{Code \& Docs:} \href{https://github.com/testingautomated-usi/fashion-mnist-c}{github.com/testingautomated-usi/fashion-mnist-c}.
\end{center}
Instead of a precompiled dataset, \textsc{Imdb-c} is released as a python library allowing to generate the dataset locally. 
This is due to copyright issues with the underlying \textsc{Imdb} dataset, but also to provide additional features such as a specifiable level of corruption severity or the support for arbitrary (non-Imdb) English text datasets.
\begin{center}
    \textbf{Install from pypi}: \texttt{pip install corrupted-text} \newline
    \textbf{Code \& Docs:} \href{https://github.com/testingautomated-usi/corrupted-text}{github.com/testingautomated-usi/corrupted-text}.
\end{center}

\paragraph{Experimental Results}
For space reasons we were only able to present a selection of aggregated results in this paper. 
The full set of results, in machine readable formats, are released as part of our \replipkg (see above).

\section{Conclusion and Future Work}
\label{sec:conclusion}

In this paper, we studied the capabilities of a wide range of different DNN test input prioritizers, along three main dimensions: (1) their capability to give high priority to misclassified (faulty) inputs; (2) their capability to identify inputs potentially useful for active learning; (3) speed. 
As a positive result, we successfully replicated the primary findings by \replicatedPaper, namely the very good performance of DeepGini, using a a more extensive and thorough experimental setup than the original paper.
On the other hand, we also found multiple alternative approaches from the uncertainty quantification literature, including the simplest ones, to perform comparably well as DeepGini. 
This finding not only strongly suggests the use of uncertainty quantifiers as TIP for DNNs,
it also strongly emphasizes the need for future work to compare against even the most naive baselines (e.g., Vanilla Softmax), and to show the advantages of novel approaches both empirically and analytically over simple uncertainty quantifiers.

Our future research on DNN test input prioritization will focus on the identification and quantification of secondary requirements for TIP, beyond early fault detection and active learning, such as diversity in the selected tests, or targeted prioritization aiming at specific root causes  for DNN faults.

\begin{acks}
This work was partially supported by the H2020 project PRECRIME,
funded under the ERC Advanced Grant 2017 Program (ERC Grant Agreement n. 787703).
\end{acks}

\balance
\bibliographystyle{ACM-Reference-Format}
\bibliography{main}
\end{document}